\pdfoutput=1

\documentclass[11pt]{article}

\usepackage[preprint]{acl}

\usepackage{times}
\usepackage{latexsym}

\usepackage[T1]{fontenc}

\usepackage[utf8]{inputenc}

\usepackage{microtype}

\usepackage{inconsolata}

\usepackage{graphicx}
\usepackage{amsmath}
\usepackage{amssymb}
\usepackage{multirow}
\usepackage{booktabs}
\usepackage{lipsum}
\usepackage[most]{tcolorbox}
\definecolor{mygreen}{RGB}{1,113,0}
\usepackage{enumitem}
\usepackage{algorithm,algpseudocode}
\usepackage{adjustbox}

\title{FReM: A Flexible Reasoning Mechanism for Balancing Quick and Slow Thinking in Long-Context Question Answering}

\author{
  Zhengyi Zhao$^{1}$, Shubo Zhang$^{2}$, Zezhong Wang$^{1}$, Bin Liang$^{1}$, \\ 
  \bf Binyang Li$^{2}$, Kam-Fai Wong$^{1}$ \\
  $^1$The Chinese University of Hong Kong,  $^2$University of International Relations\\
  \texttt{\{zyzhao, kfwong\}@se.cuhk.edu.hk}
  }

\begin{document}
\maketitle
\begin{abstract}
    Long-context question-answering (LCQA) systems have greatly benefited from the powerful reasoning capabilities of large language models (LLMs), which can be categorized into slow and quick reasoning modes. However, both modes have their limitations. Slow thinking generally leans to explore every possible reasoning path, which leads to heavy overthinking and wastes time. Quick thinking usually relies on pattern matching rather than truly understanding the query logic, which misses proper understanding. To address these issues, we propose FReM: Flexible Reasoning Mechanism, a method that adjusts reasoning depth according to the complexity of each question. Specifically, FReM leverages synthetic reference QA examples to provide an explicit chain of thought, enabling efficient handling of simple queries while allowing deeper reasoning for more complex ones. By doing so, FReM helps quick-thinking models move beyond superficial pattern matching and narrows the reasoning space for slow-thinking models to avoid unnecessary exploration. Experiments on seven QA datasets show that FReM improves reasoning accuracy and scalability, particularly for complex multihop questions, indicating its potential to advance LCQA methodologies.
\end{abstract}

\section{Introduction}

In recent years, large language models (LLMs) have demonstrated remarkable progress in long-context question answering (LCQA), achieving impressive performance across a variety of datasets \cite{caciularu2022long,liu2024lost,bai-etal-2024-citrus}. These advancements have enabled LLMs to adapt quickly to new tasks with minimal examples, significantly reducing the need for large-scale data annotation. Nevertheless, the fundamental limitations persist: either overthinking with slow thinking models \cite{yao2024hdflow,li2024happened} or underthinking with quick thinking models \cite{geva2021did,lan2022complex}. Slow-thinking models tend to apply complex reasoning to simple questions, causing delays. On the other hand, quick reasoning models heavily rely on pattern matching, failing to solve multihop QAs.

\begin{figure}[!t]
    \centering
    \includegraphics[trim={0 0 0 0},clip,width=\linewidth]{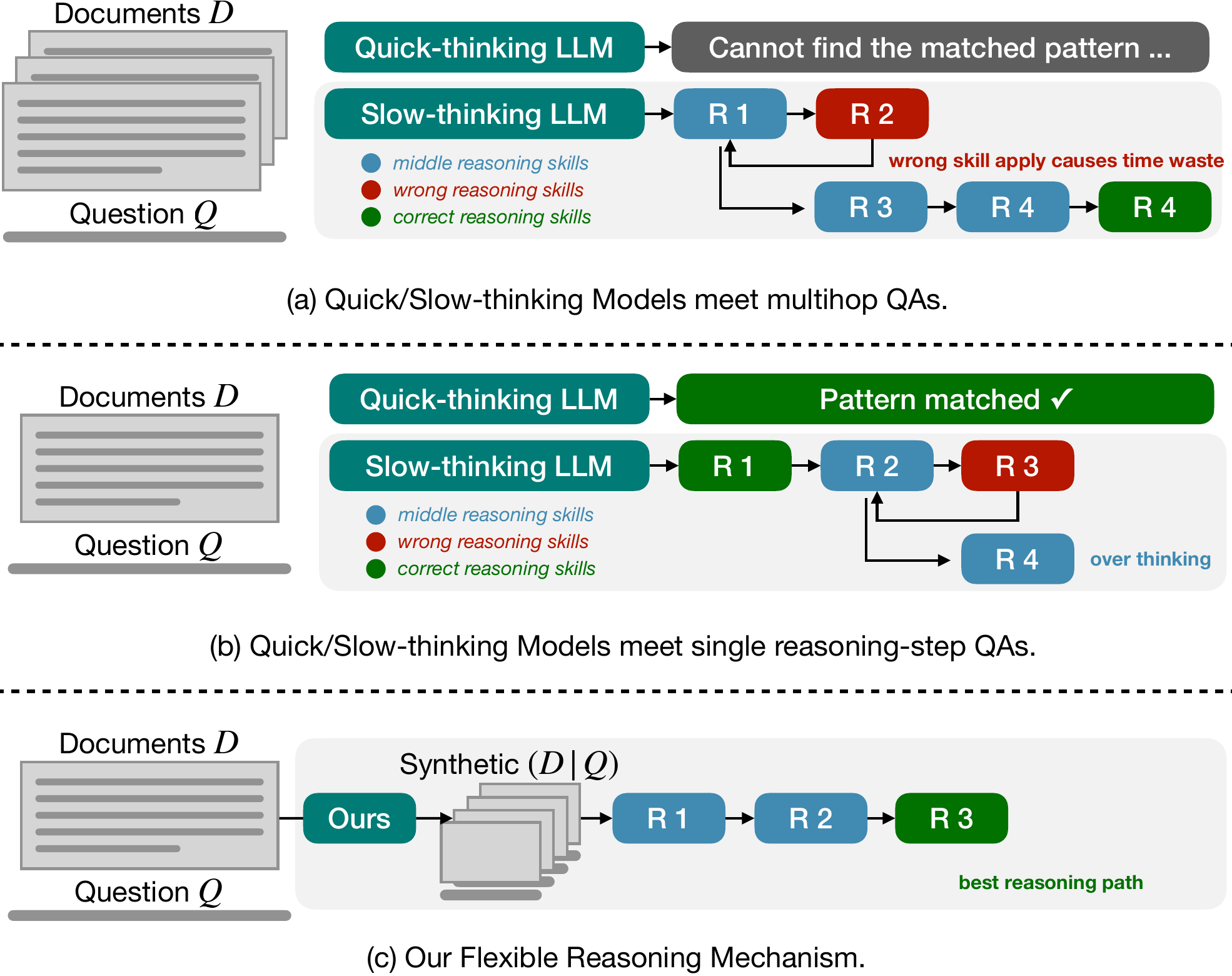}
    \caption{Demonstration of multihop reasoning in different domains. (a) shows two LLMs under multihop QA setting, (b) shows single-step QA settings. (c) shows LLMs with our proposed FReM framework.}
    \label{fig:demo}
\end{figure}

Demonstrated by Figure \ref{fig:demo}, giving a multi-hop question in (a), the quick-thinking model fails to answer a multi-hop question because it only matches patterns without deeper reasoning. The slow-thinking model attempts a long chain of thought, but it lacks a clear plan, causing errors and retracing. Then facing a single-step question in (b), the quick-thinking model can easily match the pattern to give the correct answer, but the slow-thinking model still tries to explore all possible reasoning patterns and uses the long reasoning chains to answer the question, which is unnecessary and wastes time. This raises the question: can we design a more flexible approach that balances these two thinking modes in LCQA tasks?

\begin{figure}[!t]
    \centering
    \includegraphics[width=\linewidth]{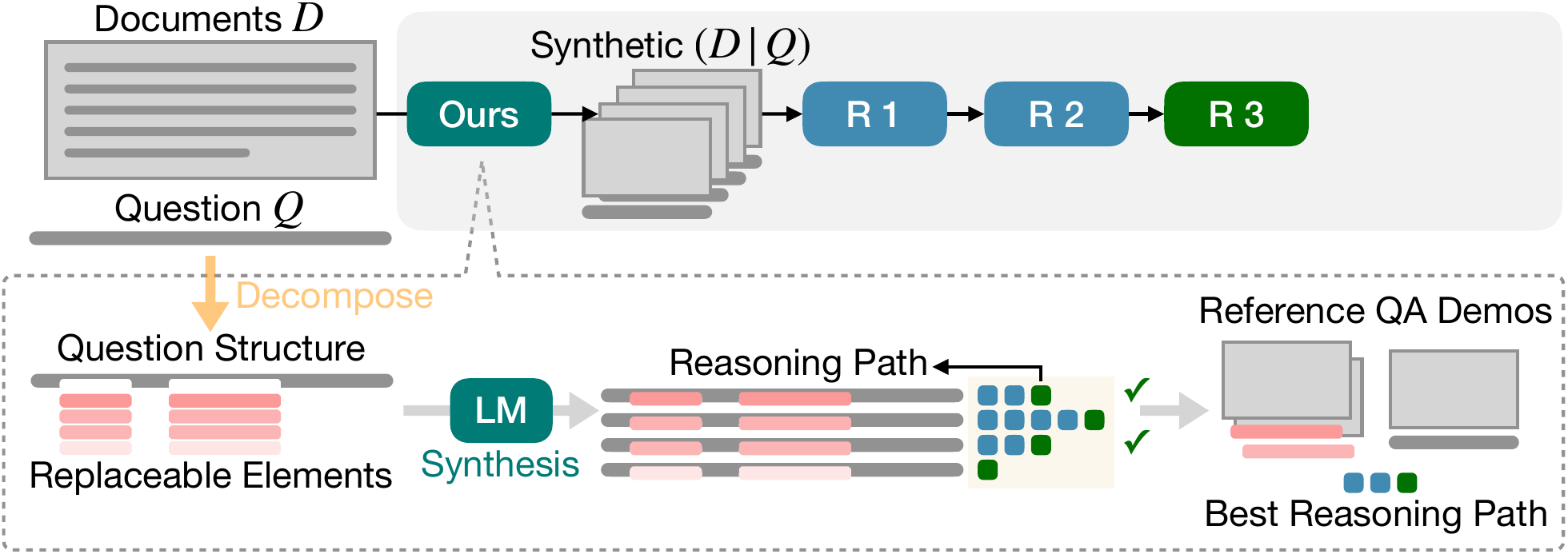}
    \caption{Overview of Reasoning-Understanding-Narrowing Mechanism. We leverage LLMs synthesis ability to explore the best reasoning path for question answering.}
    \label{fig:overview}
\end{figure}

To address these problems, we present FReM: Flexible Reasoning Mechanism. FReM tackles the inefficiency of slow-thinking approaches by narrowing the search space to curated reference demos, thereby minimizing overthinking. At the same time, it alleviates the limitations of quick-thinking methods by enforcing an explicit chain of thought, ensuring deeper logical processing rather than superficial pattern matching. Concretely, demonstrated by Figure \ref{fig:overview}, given a QA pair, FReM first decomposes the question to identify its key elements and structure. It then synthesizes multiple QA reference examples with similar question structures. Before finalizing these examples, FReM explicitly prompts the model to generate its reasoning path leveraging different reasoning skills \cite{sep-reasoning-analogy,774846}, before synthesizing the reference demos. These demos, along with their corresponding reasoning chains, are passed to a selection module, which chooses the most appropriate chain of thought for the original question. The selected chain is then used to generate the answer. In this way, FReM adapts to different question complexities: simpler queries trigger straightforward reasoning examples, while more complex questions invoke more detailed paths. By balancing these two extremes, FReM provides efficiency for simpler tasks and depth for more challenging ones.

The experiments validate the effectiveness of the proposed FReM framework across seven diverse QA datasets, including multihop and domain-specific tasks. Key results highlight FReM's ability to enhance reasoning accuracy, generalization, and scalability, particularly in complex QA scenarios.

In summary, the contributions of this work are as follows:
\begin{itemize}[leftmargin=0pt, itemsep=0pt,parsep=0pt,topsep=0pt,partopsep=0pt]
    \item We introduce a novel LCQA reasoning framework that synthesizes different reference demos to guide the model's own reasoning. This approach balances quick-thinking and slow-thinking methods by offering self-directed and flexible reasoning steps.
    \item The proposed dynamic mechanism can decide whether to use simple or more detailed reasoning pathways, depending on each question's complexity. This ensures flexible handling of both single-step and multi-hop QAs.
    \item Experimental results show our method outperforms quick-thinking models by capturing deeper logic beyond mere pattern matching. It also surpasses slow-thinking models by preventing unnecessary exploration and reducing lengthy reasoning paths.
\end{itemize}

\section{Related Work}

\paragraph{Long-Context QA.} While Question Answering (QA) tasks typically address shorter passages or single paragraphs \cite{rajpurkar2018know}, Long-context QA (LCQA) poses unique challenges because it requires models to manage and reason over extensive passages \cite{caciularu2022long,liu2024lost}. Existing works often adopt one of three strategies: (a) RAG-based retrieval \cite{zhao2024longrag}, which retrieves relevant context chunks to achieve high performance on benchmarks; (b) Iterative planning-retrieval-generation \cite{lyu2024retrieve}, which refines context and answers in multiple steps, yet can be inefficient for simple queries; and (c) Coreference resolution \cite{anonymous2025bridging}, which links entities across long passages but may be excessive for straightforward questions. However, all these methods become inefficient or overcomplicated when question complexity is low.

\paragraph{Thinking Strategies for QA.} Many QA systems operate in a ``quick-thinking'' manner by matching patterns in the input, which often breaks down on multi-hop or more nuanced questions \cite{bai-etal-2024-citrus}. In contrast, slow-thinking approaches focus on extended reasoning \cite{chen2024not}, as seen in fine-tuned long-form thought data \cite{du2025virgo} or carefully designed multi-step strategies \cite{min2024imitate,yao2024hdflow,li2024happened}. While these methods can improve performance on complex queries, they typically involve overanalysis and inefficiency for simpler tasks. Consequently, there is a growing need for more flexible approaches that adapt reasoning complexity to the actual demands of each question.

\section{FReM: Flexible Reasoning Mechanism}
\label{sec:run}

\subsection{Question Structure Identification}
\label{sec:question-structure-identification}

We have a document $D$ and a question $Q$. We split $Q$ into tokens and label those tokens that might be replaced with new terms. Then we keep the other tokens stable. We do this so our model can see which tokens in $Q$ are placeholders and which are fixed.

We let $Q = (\,q_1,\,q_2,\dots,\,q_m\,)$, where each $q_i$ is a token. We define a simple function $f_{\text{iden}}(\cdot)$ that returns $1$ if a token is a named entity or a key number, and returns $0$ if it is not. Formally,
\begin{align}
    f_{\text{iden}}(q_i) &= 
\begin{cases}
1, & \text{if $q_i$ could be replaced},\\
0, & \text{otherwise},
\end{cases} 
\\
P \;&=\;\{\,q_i \mid f_{\text{iden}}(q_i)=1\},\\
Q_{\rm S} \;&=\;\{\,q_i \mid f_{\text{iden}}(q_i)=0\}.
\end{align}
So $P$ holds placeholders, and $Q_S$ represents question structure tokens.

We label each placeholder in $P$ with a short type, like \texttt{person}, \texttt{location}, or \texttt{organization}. We form a set
\begin{equation}
T \;=\;\bigl\langle\,(p_1,\,\tau_1),\,(p_2,\,\tau_2),\,\dots,\,(p_k,\,\tau_k)\bigr\rangle,
\end{equation}
where each $p_j$ is a placeholder and each $\tau_j$ is a type. We keep $Q_S$ in its original order and merge it with placeholders from $T$. This gives us a question template with replaceable parts. For example, if $Q$ is $\text{``What is Bill Gates known for?''}$, and $q_3 = \text{``Bill Gates''}$ is recognized as a named entity of type \texttt{person}, we write $\text{``What is [entity] known for?''}$, where \texttt{[entity]} is a placeholder linked to the \texttt{person} type. 

So the final template has question structure tokens from $Q_S$ and slots from $T$. This template will guide the synthesis of related questions that share the same structure but have different placeholder content. 

\subsection{Synthetic Reference QA Generation}
\label{sec:synthetic-qa-generation}

\begin{table*}[!t]
    \centering
    \small
    \begin{tabular}{l p{0.4\textwidth} p{0.35\textwidth}}
    \toprule
    \textbf{Type of Reasoning} & \textbf{Detailed Description} & \textbf{Example} \\
    \midrule
    \textbf{Deductive} 
    & Deductive reasoning occurs when generalized statements apply to specific cases. These generalized statements are established and already proven, making specific cases easy to deduce. For example, all humans are mortals. Bill is a human, so Bill must be mortal. In this example the generalized, but proven, statement, ``all humans are mortals'' is what drives the reasoning.
    & \underline{Document:} All shapes with three sides are triangles. A certain figure here has exactly three sides.
      \newline
      \underline{Question:} What is this figure called?
      \newline
      \underline{Answer:} It is a triangle. All shapes with three sides are triangles, and this figure has three sides. So it must be a triangle. \\
    \midrule
    ......\\
    \bottomrule
    \end{tabular}
    \caption{Examples of typical reasoning steps in the chain-of-thought which typically includes multiple steps from this set of reasoning skills.}
    \label{tab:skills_demo}
\end{table*}

We now create synthetic QA pairs that share a similar question structure with the original $Q$. We use the structured template $T$ above, which has placeholders for entities or numbers. We also use a set of reasoning skills $\mathcal{S}$, such as \textit{Deductive} or \textit{Inductive} defined by \cite{sep-reasoning-analogy,774846}. Table \ref{tab:skills_demo} shows samples of defined reasoning skills.\footnote{Full definition and examples of reasoning skills are stated in Appendix \ref{apd:full_reasoning_skills}}. Each synthetic question must have a clear reasoning path made of these skills.

For each placeholder in $T$, we find possible replacements from a pool of terms that match its type by LLMselves. We do this for all $j = 1,\dots,k$, so we get a replaced question $Q_{\text{syn}}$. We combine it with the question structure tokens $Q_S$ to form a synthetic question that looks like $Q$ but uses new terms.

\noindent\textbf{Reasoning Path Specification.}
We define a sequence of reasoning path
\begin{equation}
\mathbf{s} \;=\; (\,s_1,s_2,\dots,s_L\,) \;\subseteq\; \mathcal{S}.
\end{equation}
Each $s_\ell$ is a reasoning skill, like \textit{Deductive} or \textit{Inductive}. We then synthesize a reference document $D_{\rm ref}^i$ for the specific reasoning path $\textbf{s}^i$ for all $Q^i\in Q_{\rm syn}$. We attach $D_{\rm ref}^i$ to the synthetic question $Q_{\text{syn}}^i$ to form a reference QA demo.

Then, we synthesize many reference demos by trying different replacements and different reasoning pathes $\mathbf{s}$. Let $\Gamma$ be the set of all generated demos:
\begin{equation}
\Gamma \;=\; \bigl\{ \bigl(D_{\rm ref}^i, Q_{\text{syn}}^i, A_{\text{syn}}^i, \mathbf{s}^i\bigr)\bigr\}_{i=1}^{L},
\end{equation}
where each demo has a synthetic question $Q_{\mathrm{syn}}^i$, an answer $A_{\mathrm{syn}}^i$, a reasoning path $\mathbf{s}^i = (s_1^i, s_2^i, \dots, s_{L}^i)$, and a corresponding snippets $D_{\rm ref}^i$. We store each demo as a reference QA pair with a known reasoning path. This collection helps cover simple and complex question types, so the model can later pick the most suitable path when it sees a new question.

\subsection{Selection with Multi-Criteria Matching}
\label{sec:demo-selection}

After creating a large set of synthetic QA demos. Our goal is to pick one demo that matches the original question $Q$, even if $Q$ requires unique or rarely used skills.

\paragraph{Skill Uniqueness and Coverage Computing.} We let $\mathrm{freq}(s)$ be the number of demos in $\Gamma$ that include skill $s$. When $\mathrm{freq}(s)$ is low, $s$ is considered unique. We define a uniqueness factor $\alpha(s)$:
\begin{equation}
\alpha(s) \;=\; \ln\Bigl(\tfrac{M + 1}{\mathrm{freq}(s) + 1}\Bigr),
\end{equation}
which grows when $s$ appears less often. We also define a coverage function that checks how many relevant skills from $Q$ appear in the chain $\mathbf{s}^i$. Let $\mathcal{S}_{Q}$ be a small set of skills that $Q$ is likely to need. We define
\begin{equation}
\mathrm{cover}\!\bigl(\mathbf{s}^i,\;\mathcal{S}_{Q}\bigr)\;=\;\tfrac{|\mathbf{s}^i \,\cap\, \mathcal{S}_{Q}|}{|\mathcal{S}_{Q}|}.
\end{equation}
A demo with stronger coverage addresses more of the skills demanded by $Q$.

\paragraph{LLM-Based Computing.} We also prompt a helper language model to evaluate how well each demo's question $Q_{\mathrm{syn}}^i$ aligns with $Q$. We provide $Q$ and $Q_{\mathrm{syn}}^i$ together and ask whether they share a similar structure. The response is mapped to score from 1 to 10. A higher score suggests that $Q_{\mathrm{syn}}^i$ probably resembles $Q$ in both style and complexity. We only keep demos with scores higher than threshold $\delta$.

When we have a smaller set of demos, we compute a score $W_i$ for each:
\begin{equation}
W_i \;=\; \mathrm{cover}\bigl(\mathbf{s}^i,\;\mathcal{S}_{Q}\bigr)\;+\;\sum_{\ell=1}^L \alpha\!\bigl(s_\ell^i\bigr).
\end{equation}

A demo with higher skill coverage and more unique skills has a bigger score. We then choose
\begin{equation}
i^{\ast} \;=\; \arg\max_{i}\;\{W_1, W_2, \dots\}.
\end{equation}

This leads us to the single most relevant chain for $Q$. If two demos have similar scores, we prompt the LLM again to break ties by considering the textual overlap or domain closeness.

After picking $i^\ast$, we extract $\mathbf{s}^{\,i^\ast}$ and $D_{\rm ref}^{\,i^\ast}$ as the final reasoning path. The original question $Q$, document $D$, and the chosen path are combined in a final prompt to generate the final answer. We do not tune any model parameters. Everything happens through prompts that guide the LLM to reuse existing demos. This approach enables flexible reasoning for simple or complex queries, because the selection process prioritizes both coverage and skill uniqueness. 

\subsection{Reasoning Path-Guided Answering}
\label{sec:reasoning-path-guided-answering}

After finding the most relevant and appropriate reasoning path, we now use the chosen demo index $i^*$ to guide the final answer generation. Let $\Gamma[i^*]$ store $(D_{\rm ref}^{\,i^*},Q_{\mathrm{syn}}^{\,i^*},A_{\mathrm{syn}}^{\,i^*},\mathbf{s}^{\,i^*})$. The skill sequence is
\begin{equation}
\mathbf{s}^{\,i^*} = \bigl(s_1^{\,i^*},\,s_2^{\,i^*},\dots,s_L^{\,i^*}\bigr),
\end{equation}
We locate the parts of $D$ needed for each skill $s_\ell^{\,i^*}$, combine them into $D_{\mathrm{focus}}$.

\begin{algorithm}[!h]
\caption{Reasoning Path-Guided Answering}
\label{alg:answer-gen}
\begin{algorithmic}[1]
\Require $Q, D, i^*, \Gamma$
\Ensure $A$
\State $(D_{\rm ref}^{\,i^*}, Q_{\text{syn}}^{\,i^*}, A_{\text{syn}}^{\,i^*}, \mathbf{s}^{\,i^*}) \gets \Gamma[i^*]$
\For{$\ell = 1$ to $|\mathbf{s}^{\,i^*}|$}
    \State \text{Extract needed text from $D$ for skill } $s_{\ell}^{\,i^*}$
\EndFor
\State $D_{\mathrm{focus}} \gets \bigl\{\text{all extracted text}\bigr\}$
\State $A \gets \mathrm{LLM}(\langle Q,\,D_{\mathrm{focus}},\,\mathbf{s}^{\,i^*})$
\State \textbf{return} $A$
\end{algorithmic}
\end{algorithm}

We show the answing generation in Algorithm \ref{alg:answer-gen} by combining $Q$, the skill-guided segments of $D$ (denoted $D_{\mathrm{focus}}$), and the chosen CoT $\mathbf{s}$. For instance, if the question is ``Who wrote \textit{Pride and Prejudice}?,'' the path is likely short. If the question is more demanding, like ``Which city is the birthplace of the founder of the company that built the Eiffel Tower?,'' the path usually includes abductive or inductive reasoning skills. This flexibility is possible because we align $\mathbf{s}^{\,i^*}$ with the complexity of $Q$ before building the final prompt.

We call this approach adaptive because we can pick a minimal path if $Q$ is simple, or we can pick a chain with bridging or logical steps if $Q$ is more complex. Unlike slow-thinking methods, this strategy does not force long chains on easy tasks. Unlike quick-thinking methods, it can handle multi-hop or domain-specific queries. The LLM only needs one forward pass to see the reference reasoning path and the focused text. Then it returns the final answer without extra training. This avoids domain-specific tuning and also prevents unnecessary reasoning overhead. 

\section{Experiments}





\subsection{Experimental Setups}
\label{sec:experimental-setup}

\paragraph{Datasets.} We evaluate our approach on seven QA datasets from diverse domains. \textbf{SQuAD} (general-domain questions from Wikipedia) \cite{rajpurkar-etal-2018-know}, \textbf{HotpotQA} (multihop questions spanning multiple paragraphs) \cite{yang2018hotpotqa}, \textbf{BioASQ} (biomedical queries requiring specialized knowledge) \cite{283}, \textbf{NewsQA} (news-related passages) \cite{trischler2017newsqa}, \textbf{GAOKAO} (exam-oriented dataset with academic coverage) \cite{zhang2024evaluatingperformancelargelanguage}, \textbf{HQA} (historical questions focusing on chronology and figures) \cite{HOSEN2023109245}, and \textbf{TriviaQA} (Wikipedia-based trivia) \cite{joshi-etal-2017-triviaqa}. Appendix \ref{apd:datasets} summarizes dataset sizes and domains. These tasks represent both short-answer and multihop or domain-specific QA settings.

\paragraph{Baselines and Metrics.} We compare our FReM framework against \textit{slow-thinking} and \textit{quick-thinking} baselines.  Slow-thinking approaches include: \textbf{longCoT} \cite{zhang-etal-2020-dialogpt}, \textbf{proactive} \cite{deng-etal-2023-prompting}, and \textbf{proactiveCoT} \cite{deng-etal-2023-prompting}, which generate detailed chains-of-thought for every question. Quick-thinking methods are: \textbf{few-shot prompting} and \textbf{standard prompting} \cite{deng-etal-2023-prompting}, both of which use minimal or no explicit reasoning steps. We use the standard evaluation metrics already established, Exact Match (EM) and ROUGE-L for each dataset. Each metric is consistent with prior work on its respective dataset.

\paragraph{Implementation Details.} We use four open-source LLMs (\textbf{LLaMA3} \cite{dubey2024llama}, \textbf{Qwen2.5} \cite{hui2024qwen2}, \textbf{GLM4} \cite{glm2024chatglm}, \textbf{deepseek-R1} \cite{deepseekai2025deepseekr1incentivizingreasoningcapability}) and three closed-source models (\textbf{gpt-4o} \cite{hurst2024gpt}, \textbf{gpt-o1-mini}, \textbf{gpt-o1} \cite{jaech2024openai}). For synthetic QA generation, we set a maximum output length of 4,096 tokens. When deciding which synthetic reference QA demo to use, we follow our multi-criteria matching (Section \ref{sec:demo-selection}) to pick the most relevant chain of skills. We run all experiments in Python 3.9, employing standard LLM inference APIs. Unless otherwise specified, hyperparameters stay at default values for each model. No domain-specific fine-tuning is applied, ensuring a fair and consistent comparison. Detailed implementation and all prompts can be found in Appendix \ref{apd:implementation} and \ref{apd:prompts}.

\subsection{Overall Performance}
\label{sec:overall-perf}

Table \ref{tab:main_results} shows our final experimental results on seven QA datasets. Generally, quick-thinking methods generally do well on simpler tasks like SQuAD or NewsQA, yet struggle with datasets that require domain specialization (HQA, BioASQ) or multi-hop reasoning (HotpotQA). Slow-thinking baselines often improve on these harder tasks but sometimes add unnecessary complexity for straightforward queries. Our \textbf{FReM} consistently outperforms or matches both quick-thinking and slow-thinking baselines on all seven QA datasets. On more demanding tasks like HotpotQA, HQA, and BioASQ, FReM exhibits notable gains over quick-thinking methods, confirming the value of our skill-based path selection for domain-specific and multihop questions. Even for simpler tasks like SQuAD and NewsQA, FReM secures moderate but steady improvements without forcing a full chain-of-thought. Overall, our approach surpasses nearly all prior reasoning frameworks and comes close to \textbf{gpt-o1} demonstrating the effectiveness of adaptive reasoning paths across a diverse range of QA challenges.

\begin{table*}[!t]
    \centering
    \small
    \adjustbox{max width=\textwidth}{
    \begin{tabular}{lccccccc}
        \toprule
        \multirow{2}{*}{\textbf{Models}} & \multicolumn{6}{c}{\textbf{EM}} & \textbf{ROUGE-L} \\
         & \textbf{SQuAD} & \textbf{HotpotQA} & \textbf{NewsQA} & \textbf{GAOKAO} & \textbf{HQA} & \textbf{TriviaQA} & \textbf{BioASQ} \\
        \midrule
        \multicolumn{8}{l}{\textbf{\textit{Quick-Thinking}}}\\
        \; \textbf{LLaMA3} 
            & 77.28 (-2.84) & 57.83 (-2.42) & 57.02 (-2.40) & 30.82 (-1.93) & 38.33 (-1.93) & 44.09 (-1.61) & 59.13 (-2.11) \\
        \; w/ few-shots 
            & 78.41 (-1.71) & 58.58 (-1.67) & 58.08 (-1.34) & 31.52 (-1.23) & 39.14 (-1.12) & 44.95 (-0.75) & 60.03 (-1.21) \\
        \multicolumn{8}{l}{\textbf{\textit{Slow-Thinking}}}\\
        \; w/ longCoT 
            & 77.92 (-2.20) & 58.34 (-1.91) & 57.63 (-1.79) & 31.24 (-1.51) & 39.23 (-1.03) & 44.62 (-1.08) & 59.84 (-1.40) \\
        \; w/ proactive
            & 78.53 (-1.59) & 59.05 (-1.20) & 58.25 (-1.17) & 31.83 (-0.92) & 39.52 (-0.74) & 45.13 (-0.57) & 60.43 (-0.81) \\
        \; w/ proCoT 
            & 78.80 (-1.32) & 59.44 (-0.81) & 58.73 (-0.69) & 32.12 (-0.63) & 39.74 (-0.52) & 45.23 (-0.47) & 60.83 (-0.41) \\
        \; \textbf{w/ FReM} 
            & \textbf{80.12} & \textbf{60.25} & \textbf{59.42} & \textbf{32.75} & \textbf{40.26} & \textbf{45.70} & \textbf{61.24} \\
        \midrule
        \multicolumn{8}{l}{\textbf{\textit{Quick-Thinking}}}\\
        \; \textbf{Qwen2.5} 
            & 78.02 (-3.11) & 59.24 (-2.68) & 55.53 (-2.40) & 32.24 (-1.58) & 39.13 (-1.92) & 44.64 (-1.79) & 60.23 (-1.80) \\
        \; w/ few-shots 
            & 79.13 (-2.00) & 60.11 (-1.81) & 56.33 (-1.60) & 33.04 (-0.78) & 40.08 (-0.97) & 45.56 (-0.87) & 61.03 (-1.00) \\
            \multicolumn{8}{l}{\textbf{\textit{Slow-Thinking}}}\\
        \; w/ longCoT 
            & 78.40 (-2.73) & 59.50 (-2.42) & 55.80 (-2.13) & 32.42 (-1.40) & 39.37 (-1.68) & 45.02 (-1.41) & 60.51 (-1.52)   \\
        \; w/ proactive
            & 79.32 (-1.81) & 60.27 (-1.65) & 56.63 (-1.30) & 33.18 (-0.64) & 40.28 (-0.77) & 45.75 (-0.68) & 61.15 (-0.88) \\
        \; w/ proCoT 
            & 79.63 (-1.50) & 60.73 (-1.19) & 56.83 (-1.10) & 33.52 (-0.30) & 40.54 (-0.51) & 46.05 (-0.38) & 61.58 (-0.45) \\
        \; \textbf{w/ FReM} 
            & \textbf{81.13} & \textbf{61.92} & \textbf{57.93} & \textbf{33.82} & \textbf{41.05} & \textbf{46.43} & \textbf{62.03} \\
        \midrule
        \multicolumn{8}{l}{\textbf{\textit{Quick-Thinking}}}\\
        \; \textbf{GLM4} 
            & 76.13 (-3.13) & 56.05 (-2.72) & 53.52 (-2.22) & 30.21 (-1.36) & 37.58 (-1.58) & 40.13 (-2.23) & 57.49 (-1.63) \\
        \; w/ few-shots 
            & 77.36 (-1.90) & 57.12 (-1.65) & 54.26 (-1.48) & 30.64 (-0.93) & 38.33 (-0.83) & 41.08 (-1.28) & 58.14 (-0.98) \\
            \multicolumn{8}{l}{\textbf{\textit{Slow-Thinking}}}\\
        \; w/ longCoT 
            & 76.84 (-2.42) & 56.42 (-2.35) & 53.79 (-1.95) & 30.41 (-1.16) & 37.78 (-1.38) & 40.91 (-1.45) & 57.68 (-1.44) \\
        \; w/ proactive
            & 77.67 (-1.59) & 57.04 (-1.73) & 54.49 (-1.25) & 30.96 (-0.61) & 38.62 (-0.54) & 41.74 (-0.62) & 58.42 (-0.70) \\
        \; w/ proCoT 
            & 78.11 (-1.15) & 57.34 (-1.43) & 54.73 (-1.01) & 31.26 (-0.31) & 38.82 (-0.34) & 41.96 (-0.4) & 58.61 (-0.51) \\
        \; \textbf{w/ FReM} 
            & \textbf{79.26} & \textbf{58.77} & \textbf{55.74} & \textbf{31.57} & \textbf{39.16} & \textbf{42.36} & \textbf{59.12} \\
        \midrule
        \multicolumn{8}{l}{\textbf{\textit{Quick-Thinking}}}\\
        \; \textbf{gpt-4o} 
            & 81.74 (-2.29) & 63.22 (-2.60) & 58.72 (-1.82) & 37.52 (-2.01) & 44.91 (-2.31) & 53.13 (-2.30) & 64.13 (-1.91) \\
        \; w/ few-shots 
            & 83.24 (-0.79) & 64.33 (-1.49) & 59.62 (-0.92) & 38.66 (-0.87) & 46.12 (-1.10) & 54.48 (-0.95) & 65.21 (-0.83) \\
            \multicolumn{8}{l}{\textbf{\textit{Slow-Thinking}}}\\
        \; w/ longCoT 
            & 82.67 (-1.36) & 63.52 (-2.30) & 59.03 (-1.51) & 38.02 (-1.51) & 45.50 (-1.72) & 53.94 (-1.49) & 64.55 (-1.49) \\
        \; w/ proactive
            & 83.45 (-0.58) & 64.12 (-1.70) & 59.81 (-0.73) & 38.79 (-0.74) & 46.02 (-1.20) & 54.73 (-0.70) & 65.42 (-0.62) \\
        \; w/ proCoT 
            & 83.62 (-0.41) & 64.62 (-1.20) & 60.14 (-0.40) & 39.02 (-0.51) & 46.54 (-0.68) & 55.03 (-0.40) & 65.73 (-0.31) \\
        \; \textbf{w/ FReM} 
            & \textbf{84.03} & \textbf{65.82} & \textbf{60.54} & \textbf{39.53} & \textbf{47.22} & \textbf{55.43} & \textbf{66.04} \\
        \midrule
        \; \textbf{gpt-o1-mini}
            & \textbf{85.12} & \textbf{70.45} & \textbf{62.89} & \textbf{42.74} & \textbf{49.53} & \textbf{59.17} & \textbf{68.92} \\
        \; \textbf{deepseek-R1}
            & \textbf{86.18} & \textbf{71.43} & \textbf{63.58} & \textbf{43.24} & \textbf{49.95} & \textbf{59.83} & \textbf{69.53} \\
        \; \textbf{gpt-o1}
            & \textbf{87.04} & \textbf{72.27} & \textbf{64.06} & \textbf{43.52} & \textbf{50.57} & \textbf{60.45} & \textbf{70.25} \\
        \bottomrule
    \end{tabular}
    }
    \caption{Overall experimental results on seven QA datasets. Values in parentheses show the relative difference from the ours \textbf{FReM} method. We consider gpt-o1-mini, deepseek-R1, and gpt-o1 as the default slow-thinking models.}
    \label{tab:main_results}
\end{table*}

\subsection{Effectiveness of Multihop Reasoning}
\label{sec:effectiveness-multihop}

\begin{figure}[!t]
    \centering
    \includegraphics[width=\linewidth]{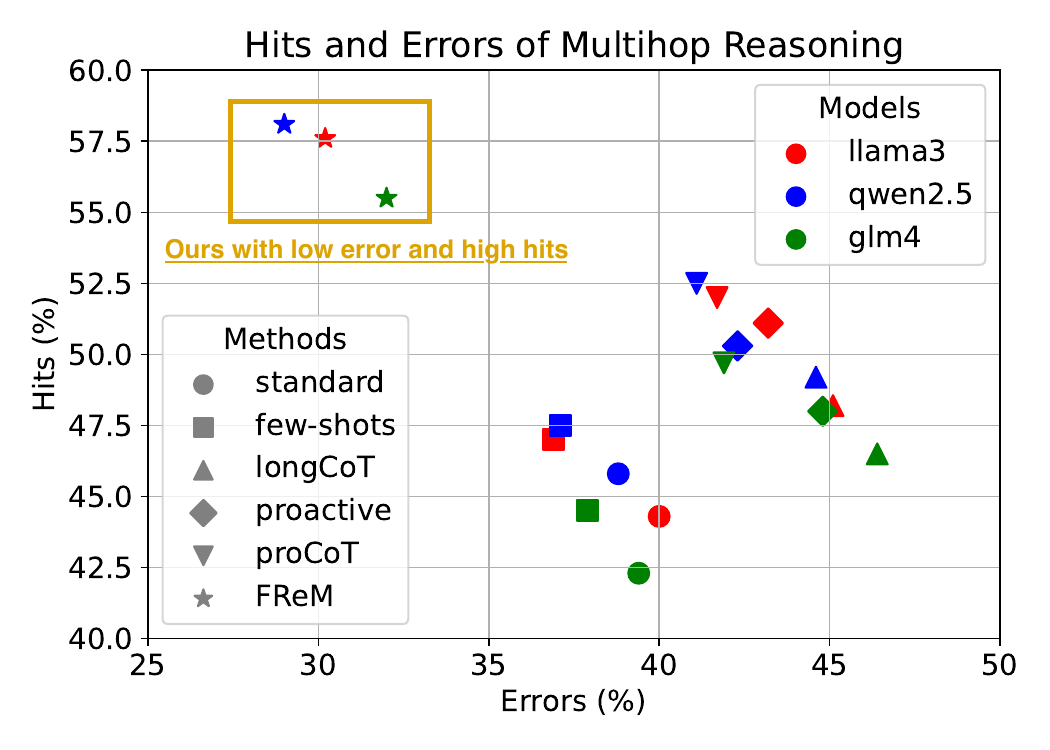}
    \caption{Extended experimental results on three open-source models for HotpotQA.}
    \label{fig:comparison_line_plots}
\end{figure}

We study HotpotQA, which requires linking multiple pieces of evidence to answer a single question. Figure \ref{fig:comparison_line_plots} shows \textit{Hits (\%)} and \textit{Errors (\%)} for five baselines and our \textbf{FReM} method on LLaMA3 backbone. Hits indicate the ability to find correct reasoning skills. Errors show how many times the model uses the wrong skill. Quick-thinking methods often fail to connect multiple clues, yielding lower hits and moderate error rates. Slow-thinking methods use detailed reasoning but can make more errors because a longer chain-of-thought amplifies the chance of going off track. Our FReM framework balances these extremes by focusing on only the necessary skill path for each question, achieving the highest EM and best hits-to-errors ratio. The experimental results on all datasets can be found in Appendix \ref{apd:multihop-reasoning}.

\subsection{Retracing and Rethinking Analysis}
\label{sec:retrace-analysis}

\begin{figure}[!t]
    \centering
    \includegraphics[width=\linewidth]{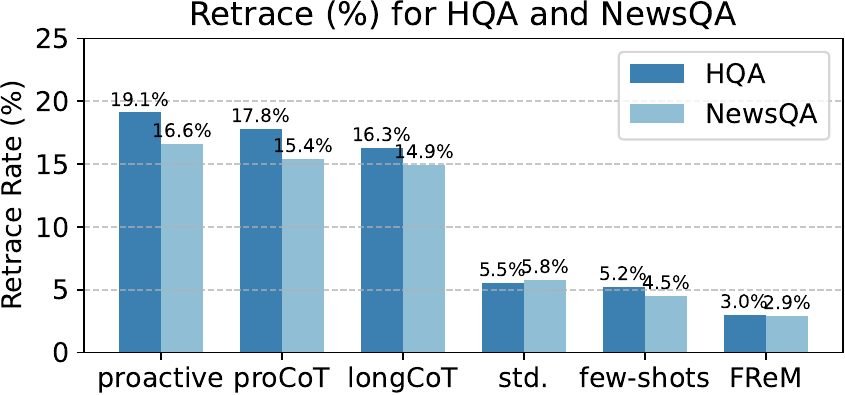}
    \caption{Retrace analysis on NewsQA and HQA.}
    \label{fig:retrace_news_hqa}
\end{figure}

We now study single-step scenarios where the question often requires only one or two direct hops. For instance, \textbf{NewsQA} focuses on straightforward articles, and \textbf{HQA} asks about historical events that typically do not demand complicated multi-hop references. Figure \ref{fig:retrace_news_hqa} reports model performance (EM) and the \textit{Retrace} rate (\%), i.e., the percentage of answers where the model rethinks or modifies its reasoning steps partway through. The experimental results on all datasets can be found in Appendix \ref{apd:singlehop-reasoning}.

Experimental results show approaches like \textit{longCoT}, \textit{proactive}, or \textit{proCoT} often generate lengthy chains-of-thought, even for single-step questions. They achieve decent accuracy but make many unnecessary inferences. This causes frequent retracing because the model must backtrack on intermediate claims that add little value. Besides, our \textbf{FReM} method performs likely to \textit{standard} or \textit{few-shots} rely on minimal or no explicit reasoning. They seldom retrace because they skip detailed steps. However, their final accuracy can suffer when even a single short reasoning hop is overlooked. Hence, Quick-Thinking obtains less overhead but might miss subtle details.

As Figure \ref{fig:retrace_news_hqa} indicates, slow-thinking methods have a notably higher retrace rate, sometimes exceeding 15\%. This happens because extra steps add confusion for shorter queries. Quick-thinking is succinct, so it sees fewer retraces but also lower accuracy. FReM's selection mechanism avoids both extremes, delivering top EM with only a small fraction of answers requiring rethinking.

\subsection{Further Analysis of FReM}
\label{sec:further-analysis}

\subsubsection{Impacts of Question Structure}
\label{sec:question-structure-impact}

Our framework decomposes each question into a \textit{structure} plus \textit{replaceable elements}. We hypothesize that questions with more placeholders benefit more from FReM's selection mechanism, because these questions allow a wider range of possible synthetic demos. Conversely, simpler questions with fewer placeholders may not need advanced reasoning paths.

We categorize questions into three buckets based on the number of placeholders in $Q$: \textit{Low} (0--1 placeholders), \textit{Medium} (2--3 placeholders), and \textit{High} (4+ placeholders). Table \ref{tab:structure_complexity} shows the performance across these groups for SQuAD and HQA to show impacts on general and domain-specific scenarios.

\begin{table}[t]
    \centering
    \small
    \begin{tabular}{lccc|ccc}
        \toprule
        & \multicolumn{3}{c}{\textbf{SQuAD}} & \multicolumn{3}{c}{\textbf{HQA}} \\
        \textbf{Group} & F-S & L-C & FReM & F-S & L-C & FReM \\
        \midrule
        Low    & 78.5 & 79.1 & \textbf{80.2} & 42.7 & 43.3 & \textbf{44.6} \\
        Medium & 76.4 & 78.2 & \textbf{79.5} & 41.5 & 43.0 & \textbf{45.1} \\
        High   & 75.9 & 78.7 & \textbf{80.1} & 40.2 & 42.9 & \textbf{46.2} \\
        \bottomrule
    \end{tabular}
    \caption{EM scores by question complexity. We compare quick-thinking (F-S indicates few-shots) and slow-thinking (L-C indicates longCoT), and our FReM.}
    \label{tab:structure_complexity}
\end{table}

As seen in Table \ref{tab:structure_complexity}, questions with more placeholders (High) see the largest gap between FReM and either baseline. This suggests that, for complex questions, enumerating and reusing relevant skill chains is particularly helpful. On simpler questions (Low), FReM still improves performance but by a smaller margin, as fewer placeholders limit the search space for alternative question structures.

\subsubsection{Synthetic Reference QA Demo Effects}
\label{sec:synthetic-demo-effects}

Our framework leverages synthetic QA demos $\Gamma$ to guide the model's reasoning on a new question. We now study how the \textit{quantity} and \textit{construction methods} of these demos affect overall performance.

\begin{figure}[!t]
    \centering
    \includegraphics[width=0.49\linewidth]{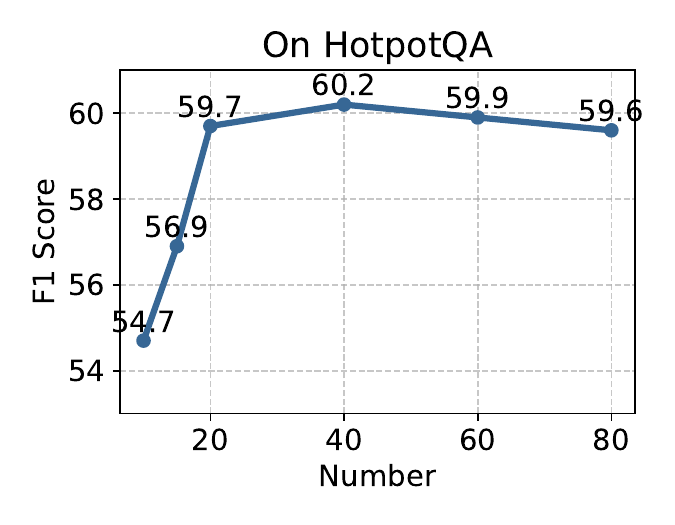}
    \hfill
    \includegraphics[width=0.49\linewidth]{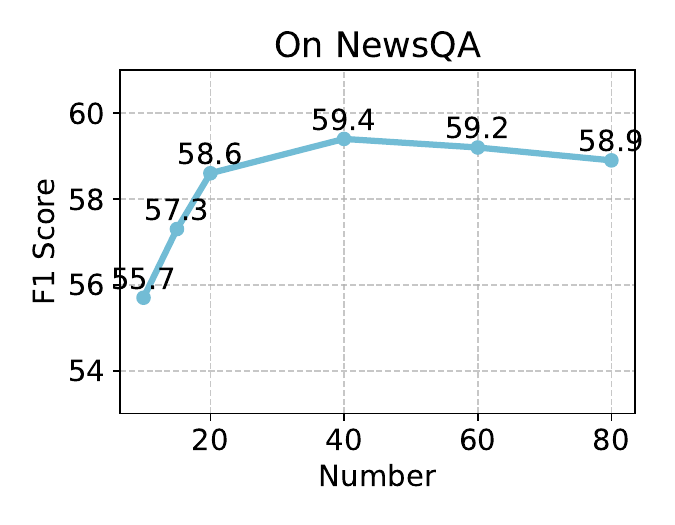}
    \caption{Impact of the number of synthetic demos on EM scores for HotpotQA (left) and NewsQA (right).}
    \label{fig:demo_counts}
\end{figure}

\paragraph{(1) Number of Demos Generated.} We vary the size $M=|\Gamma|$. Figure \ref{fig:demo_counts} illustrates the performance on HotpotQA (left) and NewsQA (right) as $M$ increases. We observe an initial boost in EM scores from $M=10$ to $M=20$, but performance plateaus or slightly decreases beyond a certain point. We hypothesize that too many demos can introduce irrelevant or redundant paths, making selection harder. In practice, we find that generating a moderate pool is enough to cover essential patterns, especially if the demos are diverse and accurate.

\begin{table}[!t]
    \centering
    \small
    \begin{tabular}{lccc}
        \toprule
        \textbf{Method} & \textbf{EM} & \textbf{Variation} & \textbf{Noise}\\
        \midrule
        \multicolumn{4}{l}{\textbf{\textit{LLaMA3 w/ ours}}}\\
        \; Random Fill            & 56.8 & High & Medium \\
        \; Guided Fill            & 58.5 & Low & Low \\
        \; Template Variation   & \textbf{59.4} & High & Low \\
        \bottomrule
    \end{tabular}
    \caption{Comparing different demo construction methods on NewsQA.}
    \label{tab:demo_construction}
\end{table}

\paragraph{(2) Demo Construction Methods.} Then, we consider how we synthesize reference demos. We experiment with different approaches for filling the placeholders:
\begin{itemize}[leftmargin=*,topsep=0pt, partopsep=0pt, itemsep=0pt, parsep=0pt]
    \item \textbf{Random Fill:} Pick random words or entities of the same type (e.g., any \texttt{person}) from a large corpus.
    \item \textbf{Guided Fill:} Use an LLM or curated list to pick semantically relevant or thematically consistent entities for each placeholder.
    \item \textbf{Template Variation:} Generate minor paraphrases or new question stems while retaining the same skill sequence.
\end{itemize}
Table \ref{tab:demo_construction} shows that template variation produces more coherent demos, with 2--4\% gains over purely random fill. This highlights the importance of a well-structured synthetic process: random replacements might yield too many off-topic or contradictory demos, while guided replacements and paraphrasing keep the demos relevant, improving the final answer selection.

\begin{figure}[!t]
    \centering
    \includegraphics[width=\linewidth]{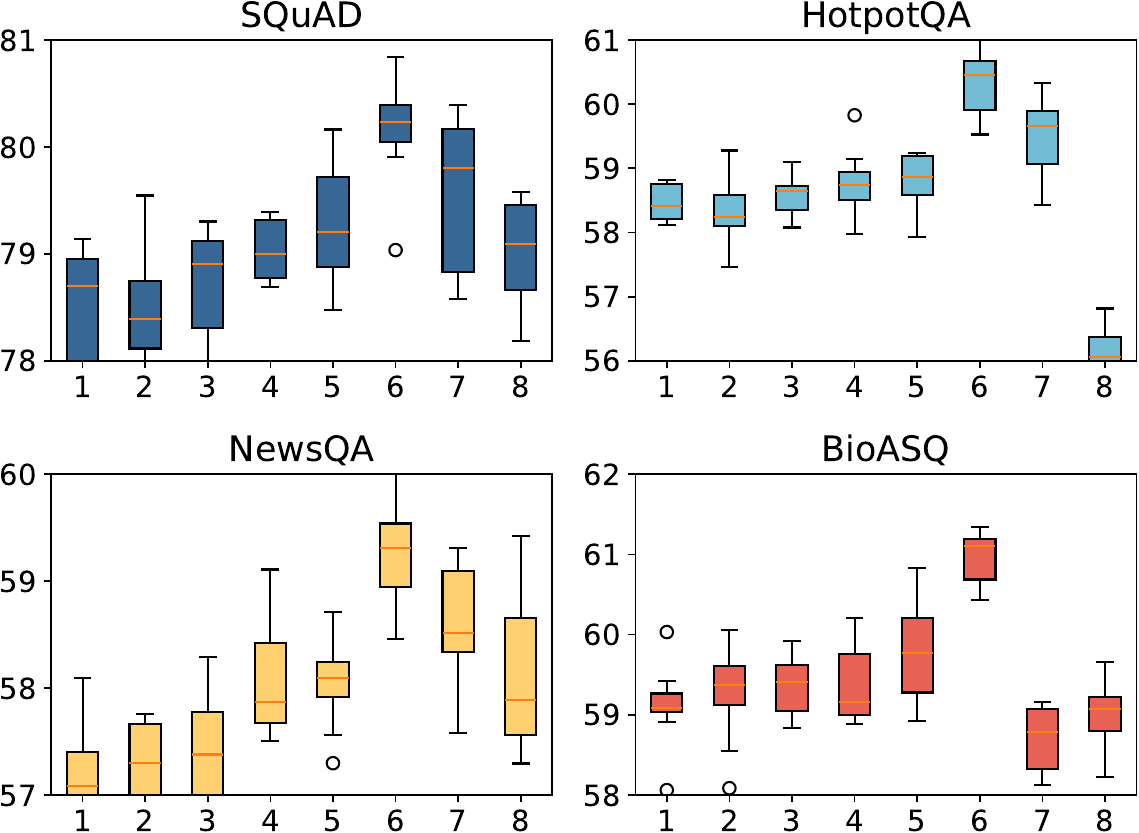}
    \caption{Impact of the question synthesis scope.}
    \label{fig:impact_of_delta}
\end{figure}

\paragraph{(3) Synthesis Question Scope under $\delta$.} We analyze the impact of varying the threshold $\delta$ on the synthesis quality of the generated questions. The threshold \(\delta\) controls how similar the synthesized questions \(Q_{\mathrm{syn}}^i\) are to the original question \(Q\), by using a helper language model to assess their alignment (in Sec.\ref{sec:demo-selection}). Figure \ref{fig:impact_of_delta} shows finding a trade-off between question similarity and generalization is much more important. As \(\delta\) increases, the similarity to the original question improves but at the cost of generalization. Conversely, when \(\delta\) is lowered, the model generalizes better but the quality of the synthesized questions decreases.

\subsection{Ablation Study}
\label{sec:ablation-study}

We conduct an ablation study to evaluate how each component influences performance. Table \ref{tab:ablation} shows removing Skill Uniqueness Weighting (SUW) causes a -5.5 drop overall, particularly lowering accuracy on HotpotQA. This indicates SUW is crucial for highlighting less-common skills. When we remove Skill Coverage Checking (SCC), performance declines more, showing that ensuring a complete set of reasoning steps is essential for both simple and multihop questions. Finally, excluding the LLM-Based Demo Judging (Judge) step yields a smaller drop, suggesting that while deleting noisy demos helps, the other two components play a more significant role in sustaining high accuracy.

\begin{table}[!t]
    \centering
    \small
    \begin{tabular}{lcccccc}
        \toprule
        \textbf{Models} & \textbf{SQuAD} & \textbf{HotpotQA} & \textbf{HQA} & $\Delta$\\
        \midrule
        \textbf{LLaMA3} & 80.1 & 60.2 & 40.2 & -- \\
        \; \textbf{w/o SUW}     & 77.9 & 58.3 & 38.9 & -5.4\\
        \; \textbf{w/o SCC}     & 78.1 & 57.8 & 38.4 & -6.2\\
        \; \textbf{w/o Judge}   & 79.4 & 59.2 & 39.6 & -2.3\\
        \bottomrule
    \end{tabular}
    \caption{Ablation results on three QA datasets with LLaMA3 as backbone model. Each row removes one component. $\,\Delta$ is the difference from Full.}
    \label{tab:ablation}
\end{table}

\subsection{Case Study}

\begin{figure}
    \centering
    \includegraphics[width=\linewidth]{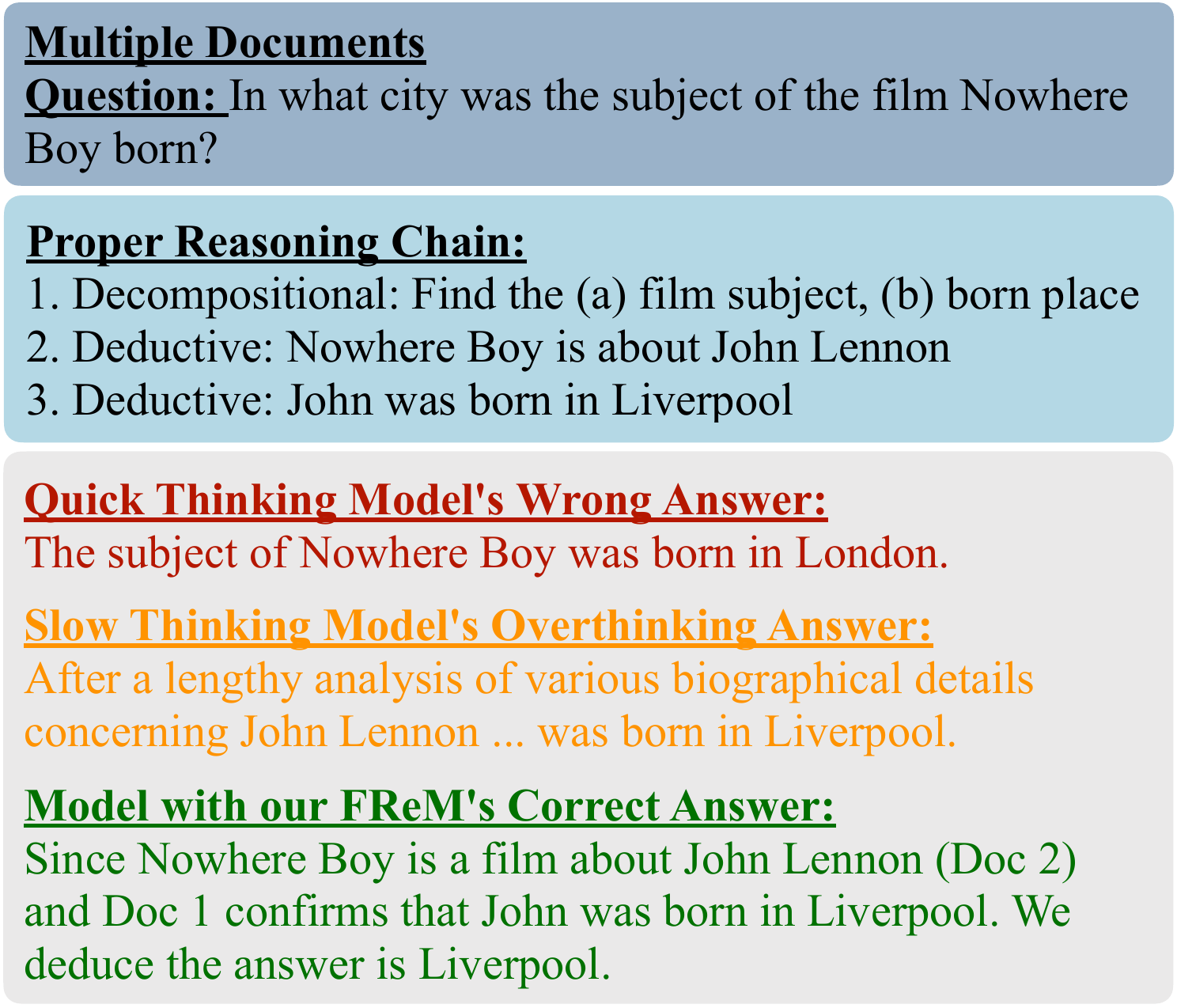}
    \caption{Case study to show effectiveness of our FReM frameworks.}
    \label{fig:case}
\end{figure}

Figure \ref{fig:case} shows the short version of example to show effectiveness of our FReM. By explicitly providing the model specific reasoning path, the model can generate the correct answer with an appropriate reasoning chain of thought. The detailed case studies can be found in Appendix \ref{apd:case}.

\section{Conclusion}

In this paper, we introduce FReM: Flexible Reasoning Mechanism, a flexible LCQA mechanism that balances quick-thinking and slow-thinking methods. By synthesizing reference demos, FReM automatically selects the most appropriate chain of thought for each question, so it neither relies on shallow pattern matching nor wastes time overthinking. Our experiments on seven QA datasets show that FReM not only promotes accuracy but also reduces unnecessary reasoning steps. Moreover, these results highlight FReM's ability to improve performance in both single-step and multi-hop scenarios, suggesting a promising direction for more efficient LCQA solutions.

\section{Limitations}

This paper did not discuss the complex effects of synthetic reference demos. Our framework uses synthetic reference demos to guide reasoning. This step needs extra preparation and careful design. If domain data is limited, generating high-quality demos becomes harder. Also, when the application domain changes, these demos may need to be updated or replaced, which adds effort. However, in the meantime, the extra preparations can also help the model to generate a better reasoning chain of thought. Future works could explore the inner effects of the reasoning chain of thoughts.

\section{Ethical Considerations}

We ensure that all experiments are conducted using publicly available, ethically sourced datasets, adhering to privacy and intellectual property guidelines. We acknowledge the potential for biases in data and are committed to evaluating and mitigating any such biases in FReM's performance. Furthermore, we emphasize the importance of model transparency, providing clear reasoning paths to enhance interpretability and accountability in decision-making. Finally, while FReM aims to improve efficiency, we are mindful of the environmental impact of large-scale model training and are committed to promoting sustainable AI practices.

\bibliography{custom}

\appendix

\section{Datasets}
\label{apd:datasets}

In this work, we evaluate our method on seven widely used question answering (QA) datasets. Each dataset presents distinct characteristics, ranging from the type of questions asked to the domain in which they are applied. Below, we provide a brief overview of each dataset.

\paragraph{SQuAD} consists of over 100,000 question-answer pairs derived from a set of Wikipedia articles. The task is to find the span of text that answers the question. SQuAD is widely used for evaluating machine reading comprehension models. The dataset includes two versions: SQuAD 1.1, which contains answerable questions, and SQuAD 2.0, which also includes unanswerable questions, making it more challenging. We use 2.0 version here.

\paragraph{HotpotQA} is a large-scale, multi-hop question answering dataset that requires reasoning across multiple supporting facts. The dataset includes over 113,000 question-answer pairs spanning various domains, where answers cannot be found in a single sentence or passage but require combining information from several documents. The questions in HotpotQA require a more complex reasoning process compared to typical single-hop datasets.

\paragraph{BioASQ} is a biomedical question answering dataset that provides information from scientific articles, primarily in the domain of biomedicine. It includes both factoid and complex questions that require understanding of scientific literature. BioASQ focuses on answering clinical, biomedical, and molecular biology-related questions using both structured and unstructured data sources.

\paragraph{NewsQA} is a dataset designed for reading comprehension tasks. It consists of over 100,000 question-answer pairs derived from news articles. The challenge of NewsQA lies in answering questions about real-world events from unstructured news stories, requiring models to handle various linguistic phenomena such as coreference, reasoning, and implicit understanding.

\paragraph{GAOKAO} is a dataset derived from the Chinese college entrance exam, also known as the "Gaokao". It contains questions related to various subjects, including Chinese literature, mathematics, and English. The questions in GAOKAO require both general knowledge and reasoning to answer. This dataset is specifically designed for the Chinese education system and is widely used in academic and educational research in China.

\paragraph{HQA} is a human-annotated dataset specifically designed for complex, open-domain question answering. It contains questions that require deep contextual understanding and can involve reasoning across long documents. The dataset includes various types of questions and answers across diverse domains, and it was created to test models' ability to perform reasoning tasks in realistic, open-ended settings.

\paragraph{TriviaQA} is a large-scale dataset that focuses on answering trivia questions, where each question is associated with a corresponding set of supporting documents. TriviaQA contains over 650,000 question-answer pairs sourced from trivia websites and requires models to retrieve relevant information from the documents and answer based on the provided facts. The dataset has questions spanning various topics such as history, geography, and general knowledge.

\section{Detailed Implementations}
\label{apd:implementation}

We use a simple pretrained language model RoBERTa for detecting named entities or key numbers in the question. This classification task involves processing the input question to identify whether it contains a named entity or key number and assigning a type to the detected entity. The model performs this task by outputting binary labels (entity: Yes/No) first, and then the associated entity types (e.g., Person, Location, Date, Organization, Number, etc.).

\begin{table}[!t]
    \centering
    \begin{tabular}{|l|l|}
    \hline
    \textbf{Parameter} & \textbf{Value} \\
    \hline
    \textbf{Model} & RoBERTa\\
    \textbf{Batch Size} & 128 \\
    \textbf{Learning Rate} & 2e-5 \\
    \textbf{Optimizer} & AdamW \\
    \textbf{Dropout Rate} & 0.1 \\
    \textbf{Evaluation Metric} & Accuracy \\
    \hline
    \end{tabular}
    \caption{Implementation parameters for named entity detection and classification.}
    \label{tab:implementation_parameters}
\end{table}

This model is fine-tuned with a simple classification layer that detects whether a named entity or key number is present in the question. The entity types are classified into categories such as Person, Location, Organization, Date, and Number. This process leverages the model's pre-trained knowledge, with minimal fine-tuning specifically focused on the entity detection and classification task.

The hyperparameters used for fine-tuning the PLM are listed in Table \ref{tab:implementation_parameters}. The batch size is set to 128, which ensures stable training and efficient model updates. The learning rate is set to \(2 \times 10^{-5}\), which is a typical value for fine-tuning transformer-based models. AdamW is used as the optimizer, as it has been shown to perform well with transformers. A dropout rate of 0.1 is applied to prevent overfitting during fine-tuning.

\section{Prompts}
\label{apd:prompts}

The primary task is to generate synthetic question-answer pairs with a reasoning path, reflecting predefined reasoning skills. Table \ref{tab:synthetic_QA_generation_prompt} shows our prompts.

\begin{table*}[!t]
    \centering
    \small
    \begin{tabular}{|l|}
    \hline
    \textbf{Prompt:} \\
    \text{You are a language model that generates synthetic question-answer (QA) pairs with reasoning paths.}\\
    \text{Your task is to generate a QA pair based on the following question. Additionally, you should}\\
    \text{provide a clear, step-by-step reasoning path that corresponds to a predefined reasoning skill.}\\
    \text{Your reasoning path should include clear substeps for each step of the thought process.} \\
    \textbf{Example 1:}\\
    \textbf{Input Question:} "Who invented the telephone?" \\
    \textbf{Step-by-step Reasoning Path:} \\
    1. Identify the key entity: "telephone" \\
    2. Identify that the question is asking for the inventor of a significant historical device \\
    3. Recall the historical context of the invention of the telephone. \\
    4. The inventor is Alexander Graham Bell. \\
    \textbf{Generated Answer:} "Alexander Graham Bell invented the telephone in 1876." \\
    \textbf{Reasoning Skill Used:} Historical knowledge of inventions. \\
    \textbf{Example 2:}\\
    \textbf{Input Question:} "What is the capital of Japan?" \\
    \textbf{Step-by-step Reasoning Path:} \\
    1. Identify the key entity: "capital" and "Japan" \\
    2. Recognize that the capital of a country is the primary city where the government is located \\
    3. Recall that Tokyo is the capital city of Japan. \\
    \textbf{Generated Answer:} "The capital of Japan is Tokyo." \\
    \textbf{Reasoning Skill Used:} Geographical knowledge of countries and capitals. \\
    \textbf{Example 3:}\\
    \textbf{Input Question:} "What is the chemical formula for water?" \\
    \textbf{Step-by-step Reasoning Path:} \\
    1. Identify the key term: "chemical formula" and "water" \\
    2. Recognize that water is a chemical compound made of hydrogen and oxygen. \\
    3. Recall the chemical composition: two hydrogen atoms and one oxygen atom. \\
    \textbf{Generated Answer:} "The chemical formula for water is H2O." \\
    \textbf{Reasoning Skill Used:} Basic chemical knowledge and scientific reasoning. \\
    \textbf{Notes:}\\
    \text{Please make sure that the reasoning path is clear and includes each substep in the thought process.}\\
    \text{The output should follow this structure: "Step-by-step reasoning," followed by the conclusion.}\\
    \text{Each reasoning skill corresponds to a specific domain of knowledge.} \\
    \hline
    \end{tabular}
    \caption{Prompt to Generate Synthetic QA Reference Demos with Reasoning Paths.}
    \label{tab:synthetic_QA_generation_prompt}
\end{table*}

Table \ref{tab:alignment_prompt} shows the helper language model for evaluating how well each demo's question aligns with original one.

\begin{table*}[!t]
\centering
\small
\begin{tabular}{|p{\textwidth}|}
\hline
\textbf{Prompt:}\\
{You are given an original question: [ORIGINAL QUESTION]}\\
{You also have a synthetic question: [SYNTHETIC QUESTION]}\\
{Your task is to decide how similar the synthetic question is in structure and complexity, compared to the original. Please provide a brief explanation of your reasoning. Then, assign a score from 1 (completely different) to 10 (very similar).}\\
\textbf{Example}:
{Original Q: "Who discovered penicillin?"}\\
{Synthetic Q: "Which scientist found the mold that led to antibiotics?"}\\
\textbf{Explanation}:
{Both questions ask about a discoverer of a major medical breakthrough. The second question focuses on the mold (penicillin), so it is structurally similar and retains the core inquiry about a discovery.}\\
\textbf{Score (1-10)}: {8}\\
\textbf{Notes}:
{- Provide a short justification.}\\
{- Avoid rewriting or changing the question.}\\
{- Keep the final output concise, ending with the numeric score.}\\
\hline
\end{tabular}
\caption{Prompt for Evaluating Alignment of Synthetic Questions with the Original.}
\label{tab:alignment_prompt}
\end{table*}

Table \ref{tab:reasoning_path_prompt} shows the question answering prompts for model.

\begin{table*}[!t]
\centering
\small
\begin{tabular}{|p{\textwidth}|}
\hline
\textbf{Prompt:}\\
{You are given:}\\
{- The original question: [Q]}\\
{- A document or context: [D]}\\
{- A selected reasoning path: [R]}\\
{- The specific skills used in the reasoning path: [S]}\\

\textbf{Your goal} is to produce a final answer by combining the relevant information from [D] with the guided reasoning steps from [R]. Follow these instructions:\\
1. \textbf{Review the Reasoning Path}\\
   Read each step in [R] carefully. Identify which parts of [D] or background knowledge support each step.  \\
2. \textbf{Apply the Skills}  \\
   If [S] includes certain reasoning skills (e.g., comparison, bridging, enumeration), make sure to explicitly use them when combining evidence from [D].\\
3. \textbf{Generate a Clear Answer}  \\
   Compose a concise final answer that directly addresses [Q]. You may outline your chain of thought, but keep the explanation aligned with [R].  \\
4. \textbf{Maintain Accuracy}  \\
   If [R] instructs a specific substep (e.g., numerical calculation or bridging multiple facts), follow it precisely, citing the relevant parts of [D].\\
\textbf{Example}:\\
{Question: "Which animal is known as the King of the Jungle?"}\\
{Document: "...the lion is often called 'King of the Jungle' due to..."}\\
{Reasoning Path:}\\
{- Step 1: Identify key subject "King of the Jungle"}\\
{- Step 2: Confirm which animal is described by that phrase}\\
{- Step 3: Conclude final answer}\\
\textbf{Answer:} "The animal known as the King of the Jungle is the lion."\\
\textbf{Notes}:  \\
- {Do not contradict the provided reasoning path.}  \\
- {Cite relevant text from [D] if needed, but avoid unnecessary repetition.}  \\
- {End with a concise, standalone final answer.}  \\
\hline
\end{tabular}
\caption{Prompt for Question Answering.}
\label{tab:reasoning_path_prompt}
\end{table*}

\section{Reasoning Skills for Reasoning Path Construction}
\label{apd:full_reasoning_skills}

Defined by \cite{sep-reasoning-analogy,774846}, reasoning can best be defined as the basic action of thinking in a sensible and rational way about something. Reasoning is the ability to assess things rationally by applying logic based on new or existing information when making a decision or solving a problem. Based on their conclusion, table \ref{tab:full_reasoning_skills} and \ref{tab:full_reasoning_skills_2} shows the reasoning skills for answering a certain question.

\begin{table*}[!t]
    \centering
    \small
    \begin{tabular}{l p{0.4\textwidth} p{0.35\textwidth}}
    \toprule
    \textbf{Type of Reasoning} & \textbf{Detailed Description} & \textbf{Example} \\
    \midrule
    \textbf{Deductive} 
    & Deductive reasoning occurs when generalized statements apply to specific cases. These generalized statements are established and already proven, making specific cases easy to deduce. For example, all humans are mortals. Bill is a human, so Bill must be mortal. In this example the generalized, but proven, statement, ``all humans are mortals'' is what drives the reasoning.
    & \underline{Document:} All shapes with three sides are triangles. A certain figure here has exactly three sides.
      \newline
      \underline{Question:} What is this figure called?
      \newline
      \underline{Answer:} It is a triangle. All shapes with three sides are triangles, and this figure has three sides. So it must be a triangle. \\
    \midrule
    \textbf{Inductive} 
    & Inductive reasoning is similar to deductive reasoning in that they both draw a conclusion based on a statement. However, in inductive reasoning, the statement is likely but has not been proven. For example, roses usually bloom in spring. In spring, one can count on there being roses. Again, the difference is that this is likely but not proven to be 100\%. 
    & \underline{Document:} Every spring for the past ten years, wild roses in Green Valley have bloomed in late March. This spring is about to begin in Green Valley.  
      \newline
      \underline{Question:} Will the wild roses bloom in late March this year?
      \newline
      \underline{Answer:} It is likely they will bloom in late March, because they usually do, but it is not guaranteed. \\
    \midrule
    \textbf{Abductive}
    & Abductive reasoning is the act of making a conclusion based on what you already know. For example, if you see a plate of food still hot, but half-eaten, you can make the conclusion that the person eating that food is probably returning soon.
    & \underline{Document:} You notice a half-eaten sandwich and a still-hot cup of coffee on a café table. The seat feels warm, and a jacket is draped over the chair.
        \newline
        \underline{Question:} Has the person who was sitting here left permanently, or are they coming back soon?
        \newline
        \underline{Answer:} It is likely they just stepped away for a moment and will return, because the food and drink are still warm and their jacket remains on the chair.\\
    \midrule
    \textbf{Cause \& Effect}
    & Cause and effect reasoning is that if x happens then y will happen as a result. This is extremely persuasive when making a speech or trying to get someone to take action to cause an effect. For example, a politician may say that if they are elected, then poverty will decrease. This is using cause and effect reasoning in a real-world situation.
    & \underline{Document:} Meteorologists predict heavy rain this evening, with warnings that streets may flood if the rainfall continues.
    \newline
    \underline{Question:} Will the roads become dangerous as a result of this weather?
    \newline
    \underline{Answer:} Yes. If heavy rain continues, roads will likely flood and become slippery, causing drivers to have less control of their vehicles.\\
    \bottomrule
    \end{tabular}
    \caption{(1/2) Full list of reasoning skills used in the reasoning path construction.}
    \label{tab:full_reasoning_skills}
\end{table*}

\begin{table*}[!t]
    \centering
    \small
    \begin{tabular}{l p{0.4\textwidth} p{0.35\textwidth}}
    \toprule
    \textbf{Type of Reasoning} & \textbf{Detailed Description} & \textbf{Example} \\
    \midrule
    \textbf{Analogical}
    & Analogical reasoning is the use of a comparison between two things to persuade that there must be more in common if they already share something. For example, if x, y, and z all share this trait, then they must also share other traits. The foundation of this type of reasoning is perfect for speeches and comparisons in the real world. If there are connections between x and y already, then they must have several other things in common as well.
    & \underline{Document:} Many leading technology companies emphasize continuous learning and adaptability. For instance, Google, Microsoft, and Amazon all invest in regular training programs and encourage innovation among employees. Their similar approach to fostering a culture of growth has been linked to their strong performance in rapidly changing markets.
    \newline
    \underline{Question:} Can we infer that a company that promotes continuous learning will also likely be successful in adapting to market changes?
    \newline
    \underline{Answer:} Yes. Since Google, Microsoft, and Amazon all share a culture of continuous learning and, as a result, demonstrate high adaptability and market success, it is reasonable to conclude by analogy that a company which also promotes continuous learning is likely to develop similar strengths.\\
    \midrule
    \textbf{Critical Thinking}
    & Critical thinking occurs when you take all of the facts and develop a conclusion based on an analysis. This could happen subconsciously or intentionally, depending on the situation. For example, in the real world, critical thinking could be about your relationships. You could see a behavior you don't like about someone and have to think critically about whether or not you will choose to spend more time with this person. This is using critical thinking to develop reasoning in a real-world application.
    & \underline{Document:} Over the past few months, Sam has repeatedly cancelled plans at the last minute and rarely communicated afterward.
    \newline
    \underline{Question:} Should you invest time in a close friendship with Sam?
    \newline
    \underline{Answer:} No. Sam's consistent behavior of last-minute cancellations suggests a pattern of unreliability, which may negatively affect the trust needed in a close friendship.\\
    \midrule
    \textbf{Decompositional}
    & Decompositional reasoning happens when the different parts of the reasoning are broken down into smaller pieces and analyzed for how they contribute to the whole. The intent of this is to make the reasoning easier to understand and allow for analyzing how the parts equal the whole. For example, in order to understand the function of the human body, you would have to analyze each bone and organ to see how they all work together. Additionally, in the real world, an argument could be broken down into several smaller parts in order to analyze the effectiveness of the argument as a whole.
    & \underline{Document:} A smartphone's quality can be understood by breaking it down into three parts: its design, performance, and battery life. The design covers the build and user interface; performance looks at processing speed and software efficiency; battery life shows how long the device operates on a single charge.
    \newline
    \underline{Question:} Can we conclude that the smartphone provides a good overall user experience?
    \newline
    \underline{Answer:} Yes. If the design is appealing, the performance is robust, and the battery life is long, then the smartphone is likely to offer a good overall experience.\\
    \bottomrule
    \end{tabular}
    \caption{(2/2) Full list of reasoning skills used in the reasoning path construction.}
    \label{tab:full_reasoning_skills_2}
\end{table*}

\section{Detailed Experimental Results}

\subsection{Further Discussion for Multihop Reasoning Scenario}
\label{apd:multihop-reasoning}

\begin{table*}[!t]
\centering
\small
\setlength{\tabcolsep}{4pt}
\begin{tabular}{lccccccc}
\toprule
\textbf{Model} & \textbf{SQuAD} & \textbf{HotpotQA} & \textbf{BioASQ} & \textbf{NewsQA} & \textbf{GAOKAO} & \textbf{HQA} & \textbf{TriviaQA}\\
\midrule
\multicolumn{8}{c}{\textbf{\textit{Hits $\uparrow$}}}\\
\textbf{\textit{LLaMA3}}\\
\; \textbf{standard}  & 63.0 & 44.3 & 23.0 & 55.4 & 68.4 & 45.1 & 32.3 \\
\; \textbf{few-shots} & 64.5 & 47.0 & 24.5 & 56.2 & 69.0 & 46.6 & 34.3 \\
\; \textbf{longCoT}   & 63.9 & 48.2 & 23.9 & 56.0 & 68.5 & 45.9 & 33.7 \\
\; \textbf{proactive} & 64.2 & 51.1 & 24.2 & 56.5 & 68.9 & 46.2 & 34.1 \\
\; \textbf{proCoT}    & 64.7 & 52.7 & 24.7 & 57.0 & 69.3 & 46.7 & 34.6 \\
\; \textbf{FReM}       & 75.4 & 57.6 & 35.4 & 59.9 & 70.2 & 67.5 & 45.8 \\
\midrule
\textbf{\textit{Qwen2.5}}\\
\; \textbf{standard}  & 59.4 & 45.8 & 24.1 & 54.2 & 65.8 & 43.0 & 31.1 \\
\; \textbf{few-shots} & 50.9 & 47.5 & 24.7 & 55.1 & 66.4 & 44.1 & 32.3 \\
\; \textbf{longCoT}   & 50.4 & 49.2 & 24.3 & 54.9 & 66.0 & 43.5 & 31.9 \\
\; \textbf{proactive} & 51.0 & 50.3 & 25.0 & 55.4 & 66.7 & 44.3 & 32.7 \\
\; \textbf{proCoT}    & 51.5 & 52.5 & 25.4 & 55.8 & 67.0 & 44.8 & 33.2 \\
\; \textbf{FReM}       & 63.7 & 58.1 & 36.7 & 55.9 & 78.1 & 55.5 & 44.6 \\
\midrule
\textbf{\textit{GLM4}}\\
\; \textbf{standard}  & 47.2 & 42.3 & 22.3 & 53.1 & 53.9 & 41.6 & 30.4 \\
\; \textbf{few-shots} & 48.7 & 44.5 & 23.0 & 53.9 & 54.5 & 42.5 & 31.4 \\
\; \textbf{longCoT}   & 48.2 & 46.5 & 22.6 & 53.7 & 54.2 & 42.1 & 30.9 \\
\; \textbf{proactive} & 48.8 & 48.0 & 23.1 & 54.2 & 54.9 & 42.7 & 31.6 \\
\; \textbf{proCoT}    & 49.3 & 49.7 & 23.5 & 54.6 & 55.2 & 42.9 & 32.1 \\
\; \textbf{FReM}       & 61.4 & 55.5 & 34.3 & 58.6 & 66.4 & 49.8 & 33.4 \\
\midrule
\multicolumn{8}{c}{\textbf{\textit{Errors $\downarrow$}}}\\
\textbf{\textit{LLaMA3}}\\
\; \textbf{standard}  & 21.2 & 40.0 & 34.6 & 38.1 & 41.9 & 44.1 & 39.3 \\
\; \textbf{few-shots} & 22.5 & 36.9 & 35.2 & 38.8 & 42.3 & 44.5 & 39.1 \\
\; \textbf{longCoT}   & 33.1 & 45.1 & 46.3 & 59.1 & 52.7 & 54.8 & 59.8 \\
\; \textbf{proactive} & 34.0 & 43.2 & 47.1 & 59.9 & 53.2 & 54.9 & 50.4 \\
\; \textbf{proCoT}    & 34.3 & 41.7 & 47.7 & 50.3 & 53.8 & 54.2 & 51.0 \\
\; \textbf{FReM}       & 18.2 & 30.2 & 29.8 & 31.5 & 34.4 & 41.8 & 34.3 \\
\midrule
\textbf{\textit{Qwen2.5}}\\
\; \textbf{standard}  & 28.9 & 38.8 & 35.0 & 36.2 & 38.8 & 40.3 & 41.7 \\
\; \textbf{few-shots} & 29.4 & 37.1 & 35.5 & 36.6 & 39.5 & 41.2 & 42.1 \\
\; \textbf{longCoT}   & 40.1 & 44.6 & 46.3 & 47.0 & 49.9 & 61.6 & 52.5 \\
\; \textbf{proactive} & 40.7 & 42.3 & 47.0 & 47.9 & 40.1 & 62.0 & 53.1 \\
\; \textbf{proCoT}    & 41.3 & 41.1 & 47.5 & 48.5 & 40.9 & 62.5 & 53.6 \\
\; \textbf{FReM}       & 28.1 & 29.0 & 30.2 & 31.7 & 22.2 & 34.3 & 24.7 \\
\midrule
\textbf{\textit{GLM4}}\\
\; \textbf{standard}  & 28.8 & 39.4 & 32.1 & 34.6 & 36.9 & 38.2 & 39.3 \\
\; \textbf{few-shots} & 29.4 & 37.9 & 32.8 & 35.2 & 37.4 & 38.6 & 40.0 \\
\; \textbf{longCoT}   & 49.9 & 46.4 & 43.2 & 45.5 & 57.9 & 43.1 & 50.4 \\
\; \textbf{proactive} & 40.3 & 44.8 & 44.5 & 46.2 & 58.3 & 45.4 & 54.2 \\
\; \textbf{proCoT}    & 40.7 & 41.9 & 45.0 & 47.1 & 59.0 & 40.0 & 51.6 \\
\; \textbf{FReM}       & 25.9 & 32.2 & 31.2 & 28.9 & 31.4 & 32.2 & 36.5 \\
\bottomrule
\end{tabular}
\caption{Full experimental results for multihop reasoning.}
\label{tab:full_results_datasets}
\end{table*}

Table \ref{tab:full_results_datasets} provides the extended experimental results on all datasets using three backbone models. For each model, we evaluate six baselines: \textit{standard}, \textit{few-shots}, \textit{longCoT}, \textit{proactive}, \textit{proCoT}, and our proposed \textbf{FReM} framework.

On general domain \textbf{SQuAD} we observe that our model significantly outperforms the quick-thinking methods in terms of hits, which indicates the correct identification of reasoning skills. While the quick-thinking models often miss context, resulting in lower hit rates, they exhibit moderate error rates. Slow-thinking models, though more precise in reasoning, often introduce errors due to overcomplicating the reasoning process. Our \textbf{FReM} framework, which focuses on synthesizing the most relevant reasoning path, strikes a balance by achieving the highest hit rate and a better hits-to-errors ratio.

Similarly, on specific domain \textbf{BioASQ}, our \textbf{FReM} method shows clear advantages. Quick-thinking models fail to connect relevant clues across the evidence, resulting in lower hit rates. On the other hand, slow-thinking models, which attempt more detailed reasoning, still face challenges as they make more errors due to the increased complexity of reasoning steps. Our \textbf{FReM} framework again excels by selecting only the necessary reasoning path for each question, yielding the best overall performance with a balance between hit rate and error reduction. This trend is consistent across other datasets further proving the effectiveness of our method in diverse QA settings.

\subsection{Further Discussion for Retracement}
\label{apd:singlehop-reasoning}

Table \ref{tab:retrace_rate} presents the retracement rates for various models across seven datasets. The retracement rate indicates the percentage of answers where the model revises or adjusts its reasoning steps during the process. A lower retracement rate is preferred as it suggests that the model is more confident in its reasoning steps and does not need to backtrack.
We observe that our \textbf{FReM} consistently achieves the lowest retracement rates across all models and datasets. The results suggest that \textbf{FReM} optimizes the balance between avoiding unnecessary retracing and maintaining high accuracy.

\begin{table*}[!t]
\centering
\small
\begin{tabular}{lcccccccc}
\toprule
\textbf{Models} & \textbf{SQuAD} & \textbf{HotpotQA} & \textbf{BioASQ} & \textbf{NewsQA} & \textbf{GAOKAO} & \textbf{HQA} & \textbf{TriviaQA} \\
\midrule
\multicolumn{8}{c}{\textbf{\textit{Retracement $\downarrow$}}}\\
\textbf{\textit{LLaMA3}}\\
\; proactive & 18.2 & 20.5 & 18.4 & 16.6 & 17.8 & 19.1 & 20.3 \\
\; proCoT    & 17.4 & 19.3 & 16.9 & 15.4 & 17.1 & 17.8 & 19.5 \\
\; longCoT   & 16.6 & 19.8 & 15.7 & 14.9 & 16.9 & 16.3 & 19.0 \\
\; standard  & 5.4  & 7.0  & 5.2 & 5.8 & 6.5 & 5.5 & 7.3 \\
\; few-shots & 3.2  & 5.2  & 4.9 & 4.5 & 4.0 & 5.2 & 5.1 \\
\; our FReM   & 2.8  & 5.2  & 3.1 & 2.9 & 3.9 & 5.1 & 4.5 \\
\midrule
\textbf{qwen2.5} \\
\; proactive & 17.8 & 19.2 & 20.2 & 15.9 & 16.5 & 18.0 & 19.4 \\
\; proCoT    & 16.9 & 18.9 & 20.5 & 14.6 & 16.2 & 16.7 & 19.2 \\
\; longCoT   & 16.3 & 19.4 & 17.0 & 14.2 & 16.1 & 15.8 & 18.7 \\
\; standard  & 5.3  & 6.8  & 6.2 & 5.7 & 6.2 & 5.4 & 7.2 \\
\; few-shots & 3.1  & 4.9  & 5.6 & 4.4 & 3.8 & 4.8 & 5.0 \\
\; our FReM   & 2.7  & 5.0  & 3.4 & 2.8 & 3.7 & 4.9 & 4.3 \\
\midrule
\textbf{GLM4} \\
\; proactive & 18.0 & 19.7 & 21.1 & 16.2 & 17.4 & 18.3 & 19.7 \\
\; proCoT    & 17.2 & 19.0 & 19.5 & 15.3 & 16.7 & 17.5 & 19.0 \\
\; longCoT   & 16.4 & 19.2 & 18.2 & 14.7 & 16.5 & 16.1 & 18.5 \\
\; standard  & 5.5 & 7.2   & 6.4 & 5.9 & 6.4 & 5.6 & 7.4 \\
\; few-shots & 3.3 & 5.0   & 5.9 & 4.3 & 4.1 & 5.0 & 5.2 \\
\; our FReM & 3.0 & 5.1 & 3.9 & 3.0 & 4.0 & 4.8 & 4.6 \\
\bottomrule
\end{tabular}
\caption{Full experimental results for retracement.}
\label{tab:retrace_rate}
\end{table*}

\section{Case Study}
\label{apd:case}

Figure \ref{fig:hotpotqa_example} and \ref{fig:squad_case_single_step} show the two different case from HotPotQA and SQuAD. The two case studies illustrate distinct reasoning strategies for question answering. In the HotpotQA case, the task requires multi-step reasoning by integrating evidence from multiple documents. A response based solely on pattern matching might output wrong ``London'' and an overthinking answer may include unnecessary details before arriving at the correct conclusion, the best approach is a concise, step-by-step explanation that clearly connects the film to John Lennon and his documented birthplace.

In contrast, the SQuAD case involves single-step reasoning where the answer is directly available from the text. The document states that the Eiffel Tower was constructed in 1889, so the correct answer is deduced by simply extracting this fact. Although a pattern matching response may correctly state ``1889'' and an overcomplicated explanation might unnecessarily discuss other numerical details, the ideal answer succinctly identifies the key fact and deduces that the construction year is 1889 using clear, deductive reasoning.

\begin{figure*}[!t]
    \small
    \begin{tcolorbox}[title=\# HotpotQA Case Example with Multihop Thinking, fonttitle=\bfseries, colframe=black, colback=white, sharp corners, boxrule=0.5mm, colbacktitle=gray!20, coltitle=black, enhanced]
        \textbf{Document 1:}\\
        John Lennon, the iconic musician and member of The Beatles, was born in Liverpool, England.\\        
        \textbf{Document 2:}\\
        \textit{Nowhere Boy} is a film that explores the early life of John Lennon, highlighting his formative years and influences.\\
        \textbf{Question:}\\
        In what city was the subject of the film \textit{Nowhere Boy} born?

        \vspace{1em}
        
        \textbf{Proper Reasoning Chain:}\\
        1. \textbf{Identify the Subject:} Recognize that \textit{Nowhere Boy} is about John Lennon. (Deductive)\\
        2. \textbf{Extract Key Fact:} From Document 1, note that John Lennon was born in Liverpool, England. (Deductive)\\
        3. \textbf{Conclude:} Therefore, the subject of the film was born in Liverpool. (Decompositional)

        \vspace{1em}
        
        \textcolor{red}{\textbf{1) Quick Thinking Model's Wrong Answer:}}\\
        \textcolor{red}{\textit{Answer:} The subject of \textit{Nowhere Boy} was born in London.}\\
        
        \textcolor{orange}{\textbf{2) Slow Thinking Model's Overthinking Answer:}}\\
        \textcolor{orange}{\textit{Answer:} After a lengthy analysis of various biographical details concerning John Lennon, including his early childhood, multiple hints in his music and interviews, and a discussion of several English cities mentioned in relation to his life, one might eventually conclude that the subject of \textit{Nowhere Boy} was born in Liverpool.}\\
        
        \textcolor{mygreen}{\textbf{3) Model with Our FReM's Correct Answer:}}\\
        \textcolor{mygreen}{\textit{Answer:} Since \textit{Nowhere Boy} is a film about John Lennon (Document 2) and Document 1 confirms that John Lennon was born in Liverpool, we deduce that the subject of the film was born in Liverpool.}
    \end{tcolorbox}
    \caption{HotpotQA Case Example with Multiple Answer Versions and Explicit Reasoning Chain.}
    \label{fig:hotpotqa_example}
\end{figure*}

\begin{figure*}[!t]
    \small
    \begin{tcolorbox}[title=\# SQuAD Case Example with Single-Step Thinking,fonttitle=\bfseries,colframe=black,colback=white,sharp corners,boxrule=0.5mm,colbacktitle=gray!20,coltitle=black,enhanced]
        \textbf{Document:}\\
        The Eiffel Tower is a wrought-iron lattice tower located on the Champ de Mars in Paris, France. It was constructed in 1889 as the entrance to the 1889 World's Fair. The tower stands approximately 324 meters tall and is one of the most recognized structures in the world.\\
        \textbf{Question:}\\
        In what year was the Eiffel Tower constructed?\\

        \vspace{1em}
        
        \textbf{Proper Reasoning Chain (using Deductive):}\\
        1) \textbf{Deduction:} Based on clear statement, we deduce that the construction year is 1889 (Deductive).
        
        \vspace{1em}
        
        \textbf{1) Quicking Thinking Model's Correct Answer \textcolor{orange}{via Pattern Matching}:}\\
        \textit{Answer:} 1889.\\
        
        \textbf{2) Slow Thinking Model's Correct Answer \textcolor{orange}{but Overthinking}:}\\
        \textit{Answer:} After reading the passage, I noticed several numbers such as 324 (the tower's height) and 1889. While I considered whether the height might play a role in determining the construction year, I ultimately realized that the passage explicitly states ``constructed in 1889.'' Thus, despite the extra details, the correct construction year is 1889.\\
        
        \textbf{3) Model with Our FReM's Correct Answer \textcolor{mygreen}{using the Specific Reasoning Chain}:}\\
        \textit{Answer:} Since the document clearly states that the Eiffel Tower was constructed in 1889, we deduce—by directly extracting and applying this fact—that the construction year is 1889.
    \end{tcolorbox}
    \caption{SQuAD Case Example with Single-Step Thinking and Multiple Answer Versions.}
    \label{fig:squad_case_single_step}
\end{figure*}

\end{document}